\documentclass[letterpaper, 10 pt, conference]{ieeeconf}  % Comment this line out if you need a4paper

\IEEEoverridecommandlockouts                              % This command is only needed if 
\usepackage{graphicx}
\usepackage{float}
\usepackage{amsmath}
\usepackage{amsfonts}
\usepackage{subfigure}
\usepackage{algorithmicx}
\usepackage{algorithm}
\usepackage{algpseudocode}
\usepackage{amssymb}
\usepackage{array}
\usepackage{booktabs}

\title{\LARGE \bf
Vision-based Robot Manipulation Learning via Human Demonstrations
}

\author{Zhixin Jia, Mengxiang Lin*, Zhixin Chen and Shibo Jian% <-this % stops a space
\thanks{*Corresponding author}% <-this % stops a space
\thanks{Z. Jia, M. Lin, Z. Chen and S. Jian are with State Key Laboratory of Software Development Envirnoment and 
        School of Mechanical Engineering and Automation, Beihang University, China.
        {\tt\small \{jiazx, linmx, zhixinc, jianshibo\}@buaa.edu.cn}}
}

\begin{document}

\maketitle
\thispagestyle{empty}
\pagestyle{empty}

\begin{abstract}
Vision-based learning methods provide promise for robots to learn complex manipulation tasks. 
However, how to generalize the learned manipulation skills to real-world interactions remains an open question. 
In this work, we study robotic manipulation skill learning from a single third-person view demonstration 
by using activity recognition and object detection in computer vision. 
To facilitate generalization across objects and environments, 
we propose to use a prior knowledge base in the form of a text corpus to infer the object to be interacted with in the context of a robot. 
We evaluate our approach in a real-world robot, using several simple and complex manipulation tasks commonly performed in daily life. 
The experimental results show that our approach achieves good generalization performance even from small amounts of training data.
\end{abstract}

\section{INTRODUCTION}
In order to accomplish tasks such as handing a cup of water, autonomous robots need to be able to perform more complex manipulation. 
Robotic manipulation is one of the central research areas in robotics, involving perception, planning, and control. 
Motion planning for complex manipulation remains a challenge, especially in unstructured dynamic real-world scenes. 
Among various existing techniques, imitation-from-observation offers considerable promise \cite{c25}\cite{c26}. 

The idea behind imitation-from-observation is to build robots that learn to behave by observing the behavior of human beings. 
Benefiting from the power of deep learning, various learning based methods work well at generating manipulation actions \cite{c27}\cite{c28}\cite{c14}. 
However, generalization to new objects and situations is still a challenge and require further effort, 
since the contexts of robot executions differ from those of human demonstrations in most practical setting. 

The difficulty with generalization lies in the fact that the manipulated objects matter a lot when performing a task. 
For example, consider a scenario in which an apple is put into a bowl by a human demonstrator.
Then, we give the robot a banana and a plate and hope that the robot could perform the similar task.
 However, the demonstrations provide incomplete information necessary to interact with new objects because they never occur before. 
 Especially, how can the robot know that it should place the banana in the plate, but not vice versa.  

 In this work, we propose a robotic manipulation skill learning approach facilitating generalization across objects and environments. 
Among all the solutions for learning from demonstration, of particular interest in this work is that only a single third-person view demonstration is used.
 This not only approximates the human way of learning behaviors, it also makes our approach salable in real world application. 
Our approach utilizes action recognition to learning manipulation behavior from human demonstration 
in the form of the sequence of actions so as to generalize across variations in context.
In addition, object detection in computer vision is applied to produce the identity and pose of an object in the context of a robot. 
 In contrast to previous methods, prior knowledge about the relationship between objects and actions is incorporated into previously 
 learned manipulation action in order to reason about the object to be interacted with. 
 Although we need collect data to train the manipulation learning model and object detecting model separately, 
 training on real or simulated robots is avoided in our method. As we known, 
 it is difficult and sometimes impossible to collect the sufficient data required by training in this situation. 
 We validate our approach through real world experiments on UR5 robot and demonstrate its robustness and generalization 
 over a wide variation in manipulation tasks and objects. 
 
The main contributions of our work are as follows. 

1) We propose a more general imitation-from-observation framework to improve generalization over objects and environments. 

2) We implement our method on an industrial UR5 robot to validate our insight and framework.

3) We further conduct experiments using manipulation tasks common in daily life to evaluate the effectiveness of robot skill generation and the generalization performance. 

The rest of this paper is organized as follows. Section II discusses the current state of robotic skill learning from visual observation. 
Section III presents the details of our approach. 
Section IV introduces system implementation in brief, followed by an experimental evaluation. Section V concludes the paper.
\section{RELATED WORKS}
\begin{figure*}
        \centering  
        \includegraphics[scale=0.82]{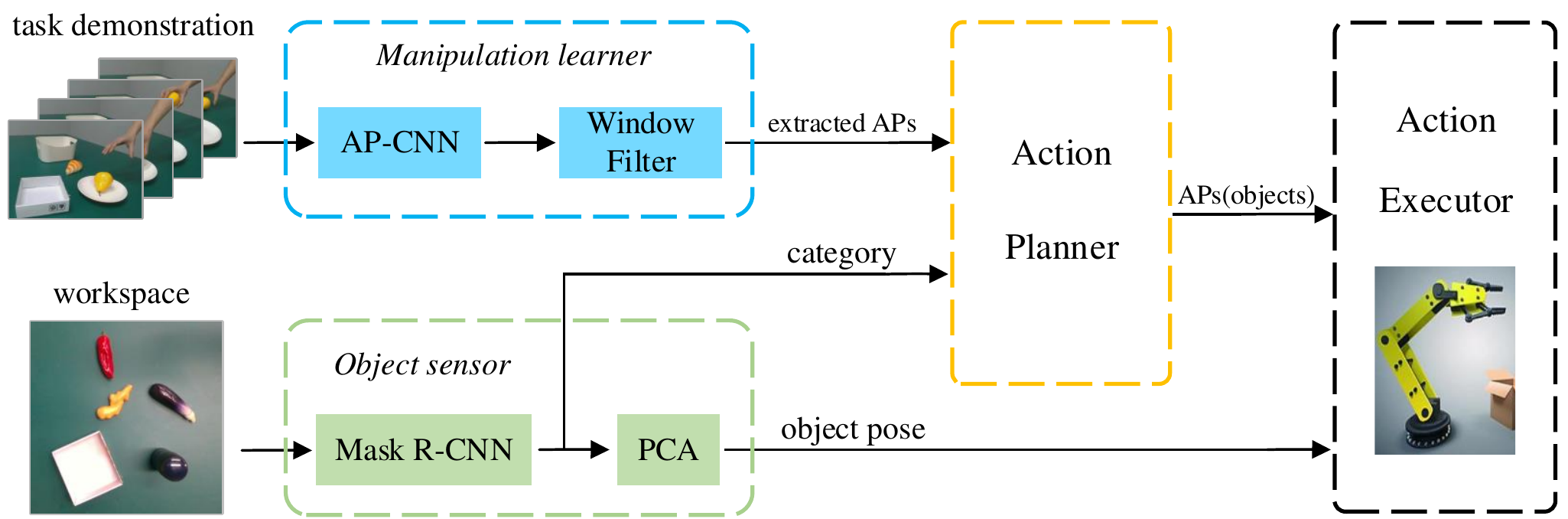} 
        \caption{The overview of our approach. In the framework, action primitive CNN(AP-CNN) is a deep convolutional neural network to fit action primitive
        classification task. Window filter is an algorithm to extract key action primitives, which is shown in Algorithm \ref{windowfilter_pseudo}.
        Mask R-CNN is used to detect and segment objects inside camera view and PCA algorithm is used to transfer the masks to 
        the physical quantities. Action planner is a Bayesian probability model based on a manually collected daily corpus, which unites actions 
        and objects for robots to execute tasks.}
        \label{overview} 
\end{figure*}
Learning from demonstration has been studied extensively in the field of robotics for decades. 
Recent advances in deep learning have made learning from observation practical, 
by which robots learn to perform manipulation tasks from observing human behavior. 
Many researchers have explored the idea of observation learning and a number of different techniques have been developed in robotics 
and artificial intelligence. For a more complete review on recent work of the field, we recommend the reader refer to \cite{c44}. 
In this section, we discuss current work related to our approach in brief.

Behavior cloning is a classical way to learn robotic manipulation, 
in which a direct mapping from states to actions is learned to reproduce either the trajectories/joint configurations of a robot, 
or a sequence of actions to be executed by a robot. The latter situation is referred to as task level learning that are concerned in this work. 
In order to acquire more information associated with human demonstration, 
new techniques developed in computer vision are investigated in the field of robot learning. 
Hejia Zhang et al. \cite{c24} propose a framework for executing collaborative manipulation tasks learned 
from full-length videos on the website based on object recognition. 
A two phased approach is introduced by Lea et al. \cite{c12}, in which features are extracted for each frame by using a Temporal Convolutional Network \cite{c13}, 
and further used for action classification. 
In addition, action recognition combined with instance segmentation also provides an effective way for robot manipulation learning. 
The work by Wang He et at. \cite{c14} propose an approach using VGG-16 \cite{c15} network to take a pair of RGB and motion images as the input, 
thereby utilizing both spatial and temporal information \cite{c13} to generate action plots for robot task execution. 
In the work by Shuo Yang \cite{c26}, a novel deep model based on grasp detection network and video captioning network is 
proposed to learn to reproduce stacking blocks and placing fruits tasks. 
Maria Koskinopoulou et al. \cite{c51} formulate a latent representation of demonstrated behaviors and associate the representation with
the corresponding robotic actions.
Cubek et al. \cite{c52} introduce a method to recognize human-demonstrated task on an object-relational abstraction layer.
Our approach does not only take use of object detection and action recognition, 
but also incorporate human knowledge into action planning. 
In this way, the generalization capability is greatly improved. 

Contrary to usual behavior cloning schemes, an alternative way is to learn perception and planning in an end-to-end fashion. 
In this setting, tasks are learned directly from raw videos of a human demonstration and a robot’s execution by using a cost function or a reward signal. 
However, this also poses a challenge for generating expected trajectories from samples in such a high dimensional input space. 
To this end, meta-learning based methods leverage demonstration data from a variety of meta-training tasks \cite{c32} \cite{c33} \cite{c25}. 
Moreover, domain-adaptive provides generalization over different objects and environments for some tasks \cite{c25}. 
In the research line inverse reinforcement learning, a lot of methods are proposed to improve the sample efficiency. 
For example, the context translation model that predicts what an expert would do in the robot context is used as reward function 
in the training of control policy \cite{c34}. The expert’s cost function is recovered with inverse reinforcement learning and then used 
to extract a policy \cite{c35}. 
Activity features obtained from the convolutional feature encoder of an activity classifier pre-trained on a large activity dataset 
are used to generate a reward signal for the learning algorithm \cite{c50}. 

\section{PROPOSED METHODS}
We present an approach that enables robots to learn vision-based manipulation skills from a third-person view demonstration.
The framework of our approach is shown in Fig. \ref{overview}, composed with four modules: manipulation learner, object sensor, action planner and action executor. 
For a given demonstration video, manipulation learner extracts the sequence of action primitives (actions in short) by using deep activity recognition models.
After that, the robot starts its manipulation task by capturing raw images produced by a camera.
Object sensor processes raw images and reasons about the identities and poses of task-relevant objects in the workspace.
To execute an action, the object being interacted with is inferred based on the current action and the likelihood distribution 
of which the object is relevant to the action. 
To this end, human knowledge about the relationship between objects and actions is introduced in action planner in the form of a text corpus. 
As a result, action planner outputs an action and the identity and pose of the object to be interacted with if necessary.
\begin{figure*}[h]
        \centering  
        \includegraphics[scale=0.76]{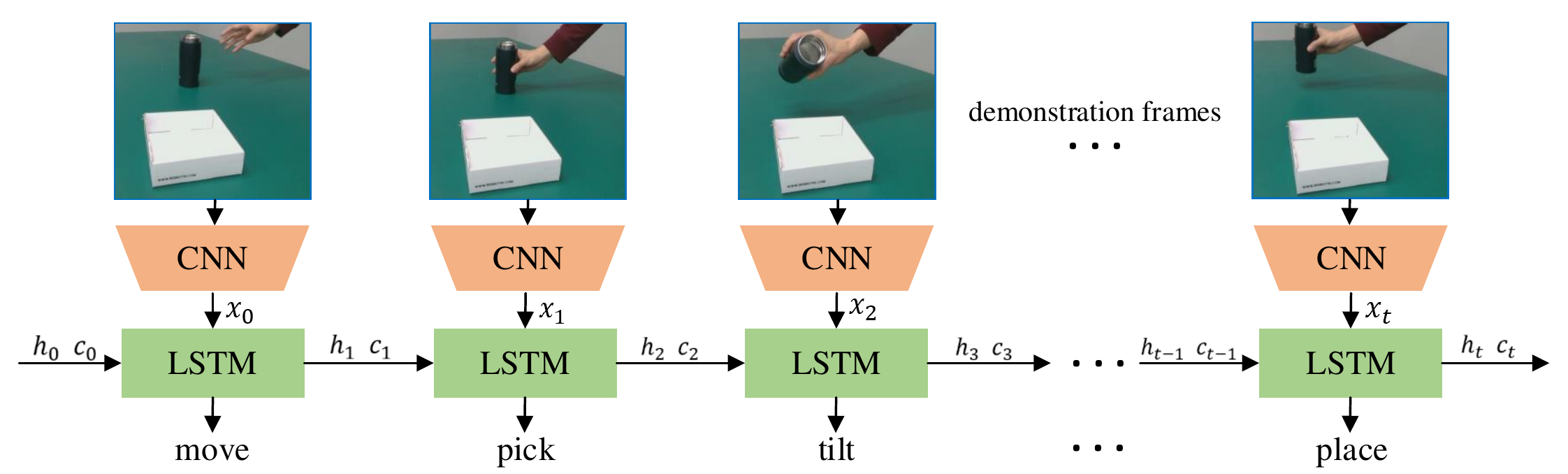} 
        \caption{The sturcture of APCNN. LSTM units are leveraged in the model to extract both temporal and spatial information.
        Each frame of the demonstration is used as input to the network. $x_i$ is the feature extracted by Resnet50 which is pretrained on Imagenet.
         $h_t$ and $c_t$ is the hidden state and cell state of LSTM cell respectively. The outputs are corresponding action primitives.} 
        \label{APCNN} 
\end{figure*}
It is worth note that actions are pre-programmed in action executor based on the basic robot operations, 
including moving and turning the end effector, and opening and closing the gripper. 
Details of our method are discussed in the following subsections.
\subsection{Manipulation Learning}
Vision-based action recognition is a powerful tool for understanding image and video in computer video. 
The basic idea of action recognition is to describe human behavior as a sequence of basic actions occurring in time order by classifying 
where an action is present or not in videos. 
In this work, we are interested in enabling robots to learn manipulation skills used in daily life, such as picking up something, 
pouring water into a cup and so on. 
Based on observation of our daily living manipulations, we define seven action primitives (\emph{idle, move, pick, place, push, tilt, rotate}), 
so as to represent human demonstration and control robot’s execution. 
For example, we could describe the manipulation task \emph{put object A into object B} as a sequence of action primitives 
\emph{$<$idle, move, pick (object A), move, place (object B)$>$}.
The executive agent of an action primitive is a human hand (end effector) and object A and B are action objects of \emph{pick} and \emph{place}, respectively. 
Note that idle means that there is no human hand (end effector) appearing in the demonstration. 

For producing a prediction of the action classes defined above, we propose an action primitive recognition network (AP-CNN), 
which consists of a CNN and a LSTM. As shown in Fig. 2, AP-CNN works in two phases: (1) extract meaningful raw image features from each frame, 
(2) using the extracted features to classify action primitive for corresponding frame. 
For feature extraction, we employ a pre-trained CNN model that takes each RGB frame of the demonstration video as input. 
Since frame-base action classification may suffer from over segmentation and the lack of local temporal coherence, 
we feed the extracted features into a LSTM network to optimize the model performance by utilizing temporal information.
The LSTM part takes the features extracted by pre-trained CNN as input and constantly updates the hidden states and cell states 
which go through the network before. 
With the help of LSTM mechanism, the network can predict the result of current frame based on previous temporal information.
Dropout layers are added to avoid overfitting at the same time.
For a given demonstration, AP-CNN generates n action primitives, where n is the number of frames in video. 
To reduce the redundancy in consecutive frames, a window based filter is applied. 
With a window moving forward from the head to the end of action primitive sequence, 
continuously identical primitives are considered as a single action primitive. 
The specific procedure of window filter is shown in Algorithm 1. 
Five different tasks and their predicted results are shown in Fig.3.

\begin{algorithm}
        \caption{Action Primitive Window Filter}
        \label{windowfilter_pseudo}
        \begin{algorithmic}[1]
        \Require Action primitive sequence, $S_n$; Window width, $w$
        \Ensure Key action primitives, $K$
        \State $i \gets 0$
        \State $K \gets \varnothing$
        \Repeat
        \State find the most frequent $S_M$ in $\left \{ S_i, S_{i+1}, ... , S_{i+w} \right \}$
        \If {$S_M$ is not the last one in $K$}
        \State append $S_M$ to $K$
        \EndIf
        \State $i \gets i+1$
        \Until{$i == n-w$}
        \State \Return $K$
        \end{algorithmic}
        \end{algorithm}

\begin{figure}[h] 
        \centering 
        \subfigure[]{ 
                \includegraphics[scale=0.435]{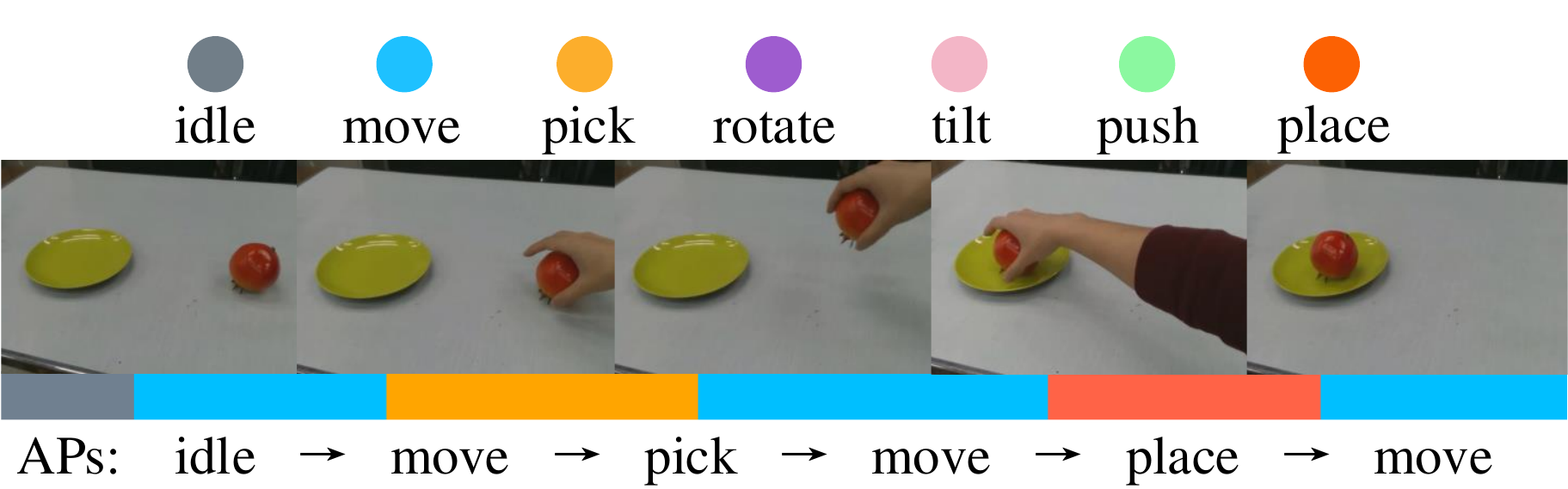}
                \label{singletask1}
        }
        \quad
        \subfigure[]{
                \includegraphics[scale=0.435]{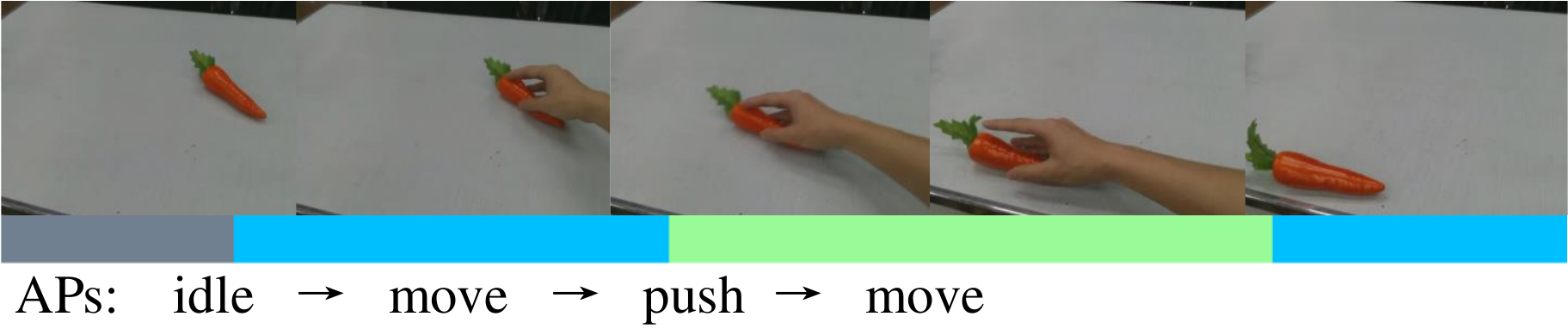}
                \label{singletask2}
        }
        \quad
        \subfigure[]{ 
                \includegraphics[scale=0.435]{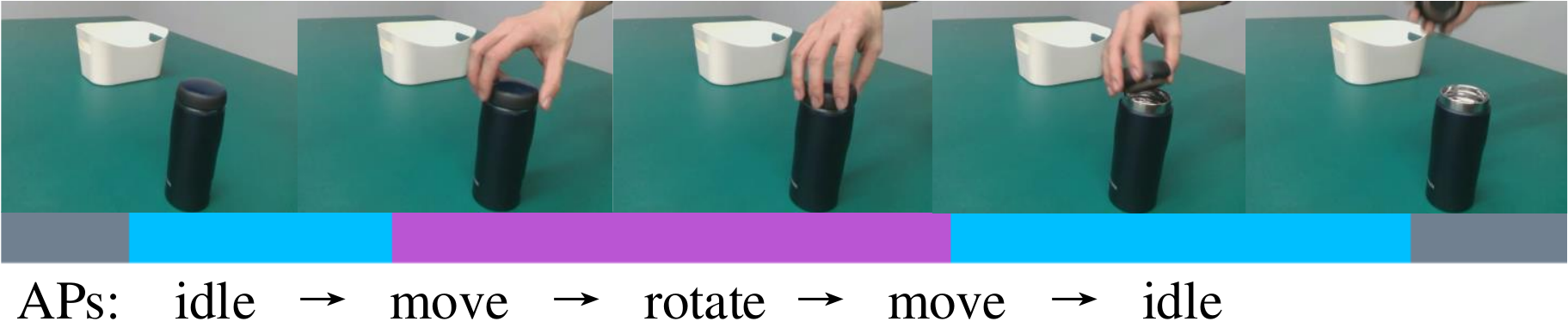}
                \label{singletask3}
        }
        \quad
        \subfigure[]{ 
                \includegraphics[scale=0.435]{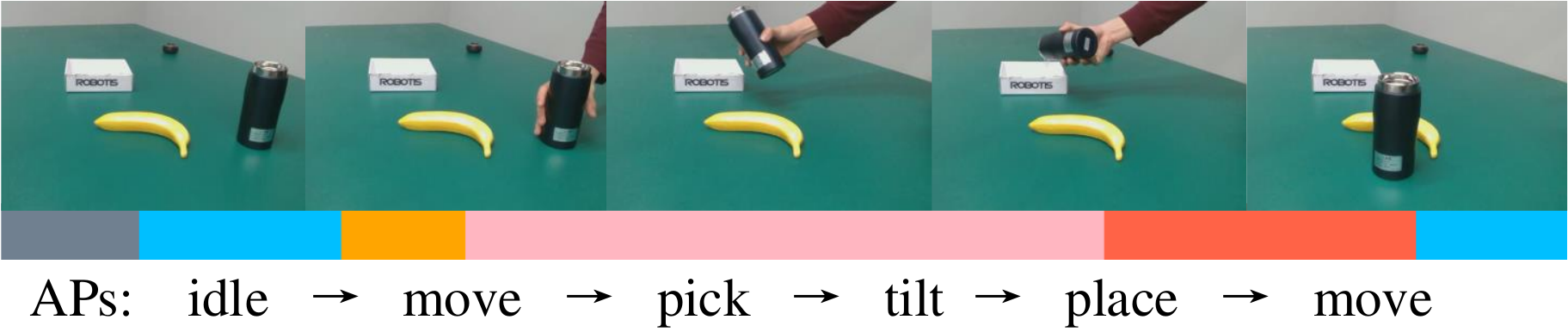}
                \label{singletask4}
        }
        \quad
        \subfigure[]{ 
                \includegraphics[scale=0.435]{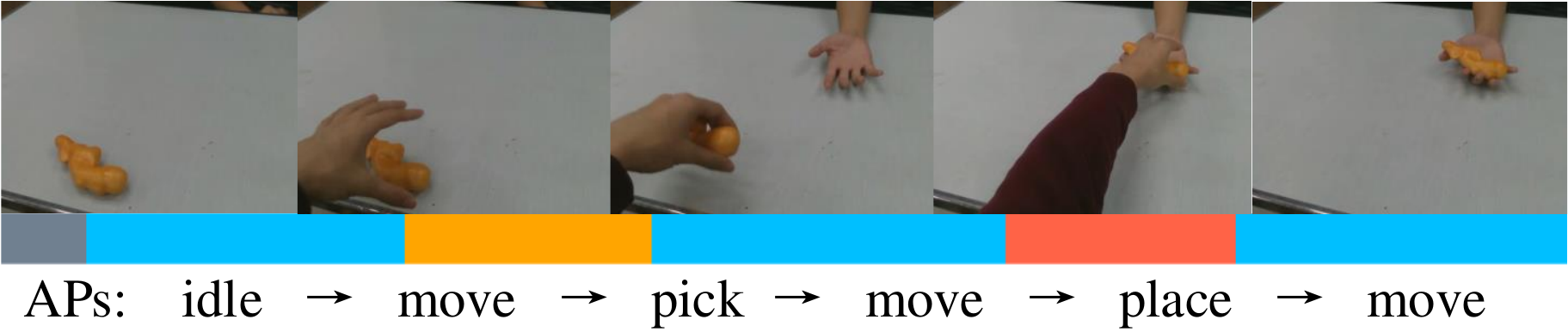}
                \label{singletask5}
        }
        \quad
        \caption{The five tasks and their corresponding action primitives recognized. 
                Action primitives are colored as shown in the first row. 
                The color bar represents the different action primitives at different time of the video. 
                (a) a pick-place task.
                (b) a pushing away task.
                (c) an openning bottle task.
                (d) a pouring task.
                (e) a delivering task.}
        \label{ap-result} 
\end{figure}
\subsection{Object Sensing}
The object sensor module plays a sensor-like role in our framework, 
through which object information including object category and object pose is computed from raw RGB images obtained by a robot camera. 
It’s worth noting that the workspace of a robot is constrained to a tabletop in this work. 
In the setting, we would estimate 3d object pose instead of 6d. 
Constrained workspace could allow us to focus on more important aspects of our framework. 
 
As shown in Fig. \ref{overview}, a popular object detection deep network Mask R-CNN \cite{c20} is applied to detect the category of objects and their corresponding masks 
from raw RGB image input. It works in two steps. 
The first is to recognize candidate object bounding boxes using Region Proposal Network (RPN) \cite{c19}. 
Then it is to predict the class, box offset and a binary mask for each region of interest (RoI). 
For presentation purposes, the result of Mask R-CNN is defined as follow:

\begin{equation}
        C = \left\{c_1, c_2, ..., c_n\right\}
\end{equation}
\begin{equation}
        M = \left\{m_1, m_2, ..., m_n\right\}
\end{equation}  
where $C$ is the set of types of objects recognized, $M$ is the set of masks on the detected objects, and $n$ indicates how many objects are recognized in total. 
Furthermore, the set of pixel points $m_i$ is expressed as follow, representing the mask that covers the corresponding object. 
\begin{equation}
        m_i = \left\{\left(x_1, y_1\right), \left(x_2, y_2\right), ... , \left(x_j, y_j\right)\right\}
\end{equation}
where $j$ is the number of pixel points in mask $m_i$. Therefore, for each detected object, 3d object pose is represented as follow:
\begin{figure} 
        \centering  
        \subfigure[]{ 
                \includegraphics[scale=0.5]{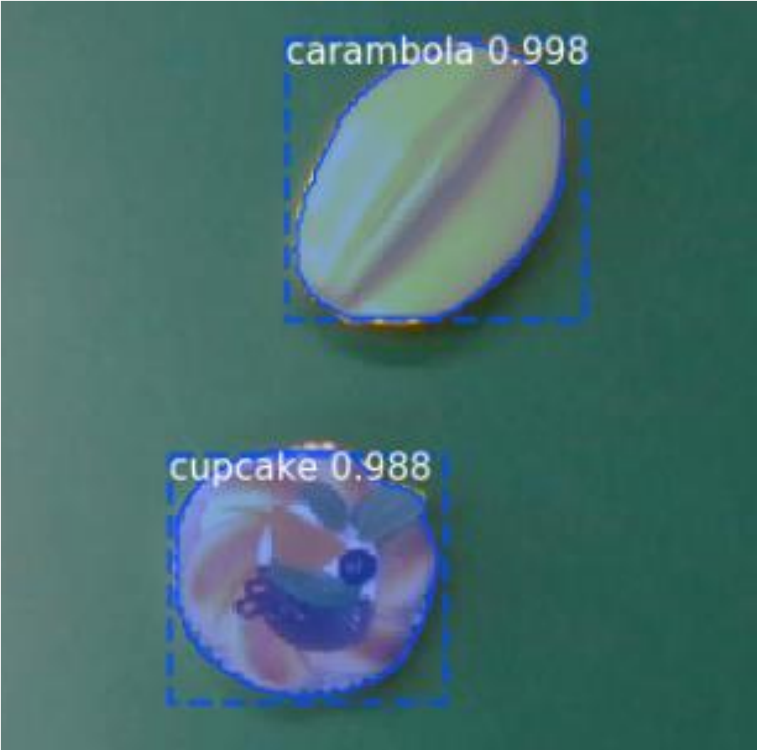}
                \label{maskresult}
        }
        \quad
        \subfigure[]{
                \includegraphics[scale=0.5]{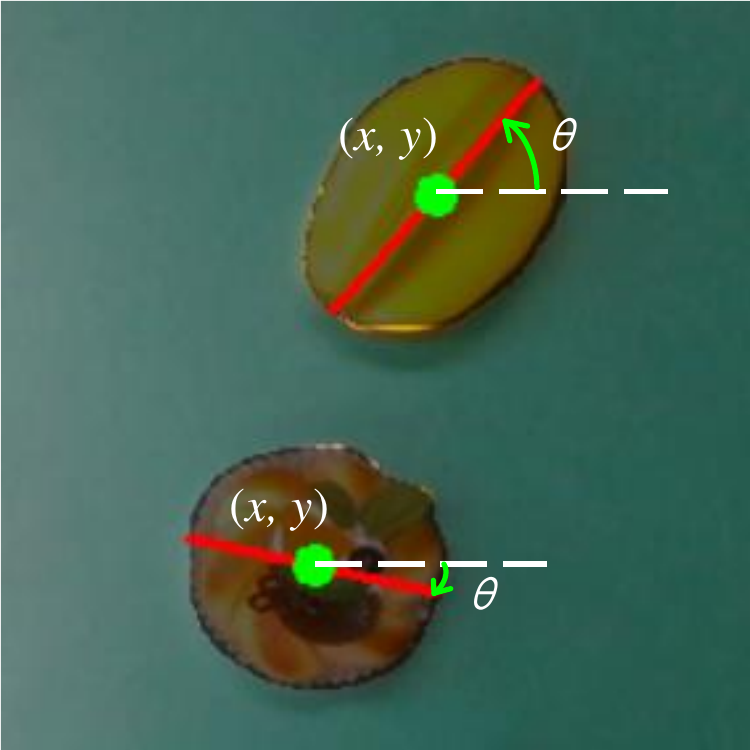}
                \label{pcaresult}
        }
        \caption{The result of Mask R-CNN is shown in (a). Blue mask covers the detected objects and the classes of objects 
        is on the top left of bounding boxes. 
        (b) shows the robotic manipulating information representation. Green point $(x,y)$ corresponds to the center of the detected object. 
        The red lines indicate the main component of the masks. $\theta$ is the orientation of the mask relative to the horizontal axis.} %图题
        \label{method-a-result} 
\end{figure}
\begin{equation}
        m = \left\{x,y,\theta,c\right\}
\end{equation}
where, $(x, y)$ is the center of a detected object, $\theta$ is the orientation of the mask relative to the horizontal axis, 
and $c$ denotes the object class. 
To estimate the object pose, we utilize a principle component analysis (PCA) based algorithm, in which, $(x, y)$ is the mean point of $n$ pixel points 
in $m_i$ and $\theta$ is the main direction with the largest variance. The example results of Mask R-CNN and the PCA are visualized 
in Fig. \ref{method-a-result}.

\subsection{Action Planning}
Action planner is an intelligent part in our framework for adapting to changing environments.
In this work, we consider the situation where a robot works under the circumstances with a varied configuration, 
instead of a constrained one. It means that robots have to plan and execute tasks in new scenes that are different from those in the human demonstration. 
To achieve this, both world knowledge and some reasoning ability are required to match object instances to the sequence of actions learned. 

To incorporate our prior knowledge about action, we construct a knowledge base in the form of text corpus. 
Our knowledge base is made up of sentences in English which describe how human users act for a given object. 
For example, pick the apple or push the pear to the white plate and etc. 
In total, we manually collect 1340 sentences according to the tasks in our experiment. 
However, it is also possible to collect sentences through learning method such as video caption. 
Theoretically, the ability of action planner can be greatly improved with more knowledge about human behavior.

Furthermore, we refer the work \cite{c24} to combine objects with actions using a probability model as follow:
\begin{equation}
        o = \arg\max_{o_i\in \mathbb{O}} P\left(o_i|A=a\right)
\end{equation}
where $\mathbb{O}$ is the detected objects set from Mask R-CNN, $a$ is the current planning action primitive from the task.
Based on the corpus, we firstly filter all the sentences related with the current action primitive $a$
and then find the object $o$ with the highest probability in these filtered sentences.
Depending on the interaction property of an action, the number of objects that action planner generates varies. 
For example, pick, place and rotate are actions involving one manipulation object, and push and tilt are related with two objects. 
In the case of one manipulation object, the object in the detected set with the highest frequency is chosen the target object. 
In other cases, all the objects in the knowledge base are sorted by their frequency and checked in order if these objects are in the detected set.

\section{EXPERIMENTS}
We implement our framework on a real world robot for empirical validation by studying the effectiveness of human demonstration based 
manipulation learning on vision-based tasks. 
In particular, five simple tasks and two complex tasks are designed to evaluate the ability to learn manipulation skills and generalization 
over objects and situations. Each trial starts with the demonstrator performing a specified task. After observing human behavior, 
the robot is to adapt manipulation learned to the objects in its own context. A test is successful if the robot performs equally as well as demonstrator. 
The video results are available at: \textbf{https://youtu.be/2zDHKNMdOFo}.

\subsection{Implement Details}
The AP-CNN is implemented on top of Resnet50 \cite{c29}, a deep network pre-trained on Imagenet for object recognition. 
We reuse the convolutional part of Resnet50, which takes a $240\times360$ images with three channels as input and outputs a 2048 dimension vector to the LSTM.
 AP-CNN ends with fully connected layer interleaved with ReLU and a softmax layer for action classification.

To collect data for training AP-CNN, we record human demonstrations involving not only simple manipulation tasks but also complex tasks. 
Specifically, 100 demonstration videos are made for each simple task, and 50 videos for each complex task. 
Finally, 600 videos in our dataset are randomly partitioned into a training set with 90$\%$
 of the data, and a test set with the remaining data. Adam optimization \cite{c23} with a learning rate of 1e-4 is used to train AP-CNN. 
 It took about 11 hours to train the AP-CNN on a single desktop PC with 64 GB RAM and a RTX 2080ti GPU. 

The implementation of Mask R-CNN is based on \cite{c20}. 
To train Mask R-CNN, we collect 1782 images involving 28 different kinds of objects taken by the RealSense \cite{c21} camera and resize the RGB images
 in the dataset into $600\times600$. All these images are labeled manually by LabelMe \cite{c22}. 

  Our hardware platform is an industrial UR5 robot arm equipped with an RG2 gripper. 
  The RealSense camera is located 100 cm above the work surface, producing images at a resolution of $600\times600$ pixels. 
  A laptop with a RTX 2080 GPU acceleration is used for real-time robotic control and communication with UR5 via TCP/IP protocol. 

\begin{figure}
        \centering  
        \subfigure[]{
                \includegraphics[scale=0.88]{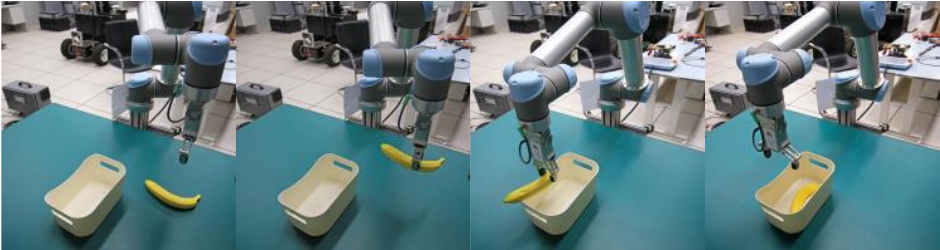}
                \label{singletasks-1}
        }
        \quad
        \subfigure[]{ 
                \includegraphics[scale=0.88]{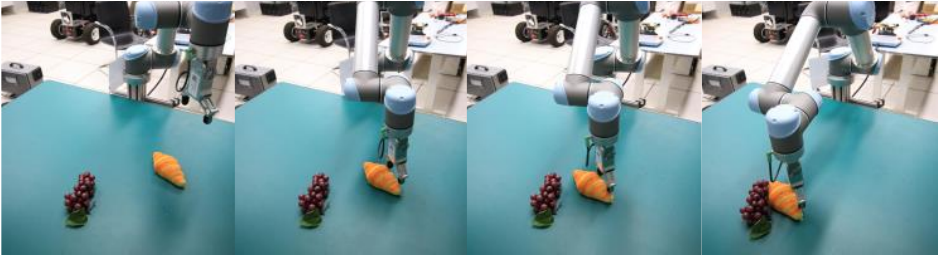}
                \label{singletasks-2}
        }
        \quad
        \subfigure[]{ 
                \includegraphics[scale=0.88]{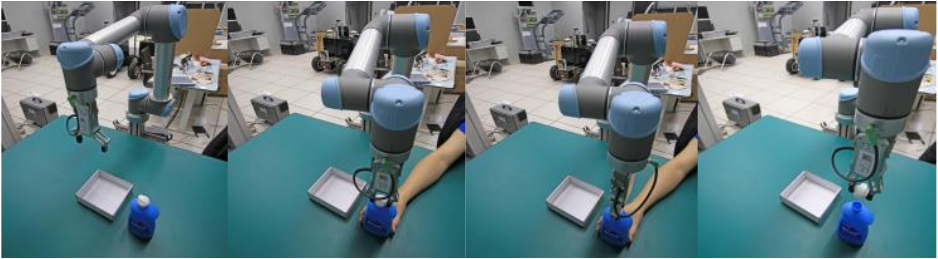}
                \label{singletasks-3}
        }
        \quad
        \subfigure[]{ 
                \includegraphics[scale=0.88]{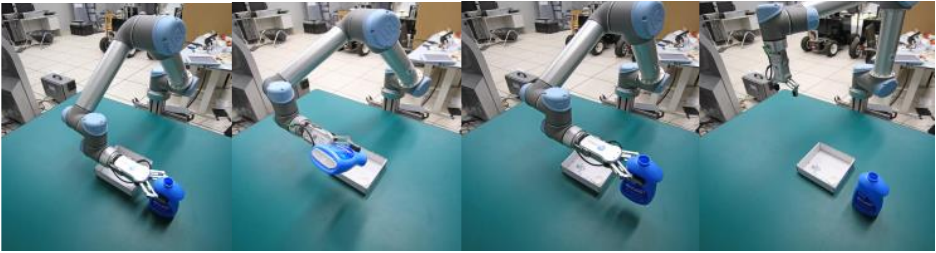}
                \label{singletasks-4}
        }
        \quad
        \subfigure[]{
                \includegraphics[scale=0.88]{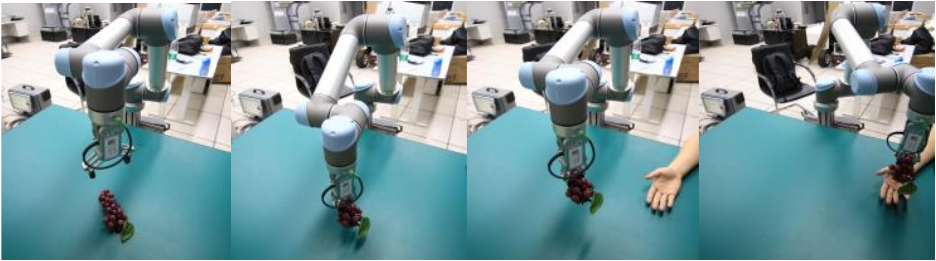}
                \label{singletasks-5}
        }
        \caption{The example results of robot learning from  demonstrations. 
        (a): a pick-place task executed by the robot, learned from Fig. \ref{singletask1}.
        (b): a pushing task executed by robot, learned from Fig. \ref{singletask2}.
        (c): an openning bottle task learnt from Fig. \ref{singletask3}.
        (d): a pouring bottle task learnt from Fig. \ref{singletask4}.
        (e): a delivering task learnt from the demonstration in Fig. \ref{singletask5}.
        Note that the objects that the robot faces when executing tasks are different from those in the demonstrations.
        } 
        \label{singletasks} 
\end{figure}
\subsection{Simple Task Learning}
We first evaluate our approach on simple manipulation tasks. To this end, the following five operations commonly used in our daily life are chosen. 

$\bullet$ Pick-place task: picking up an object and placing it into a container.

$\bullet$ Pushing task: pushing an object toward a goal position. 

$\bullet$ Opening task: turning the cap to open an object. 

$\bullet$ Pouring task: grasping an object and tilting it to pour something into a container.

$\bullet$ Delivering task: picking up an object and handing it to someone.  

The video pictures of each operation performed by human are shown in Fig. \ref{ap-result}, respectively. 
\begin{figure*} 
        \centering  
        \subfigure[]{ 
                \includegraphics[scale=1.51]{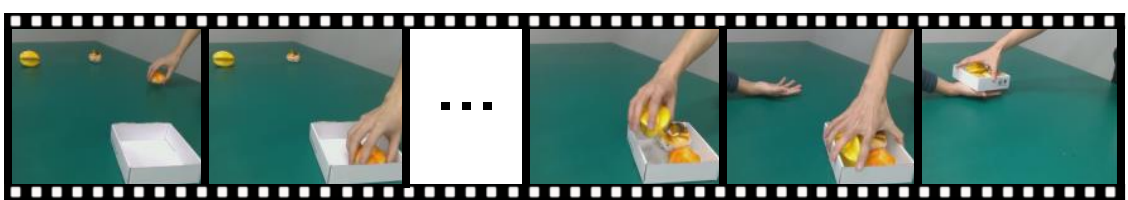}
                \label{multitask1}
        }
        \quad
        \subfigure[]{
                \includegraphics[scale=1.51]{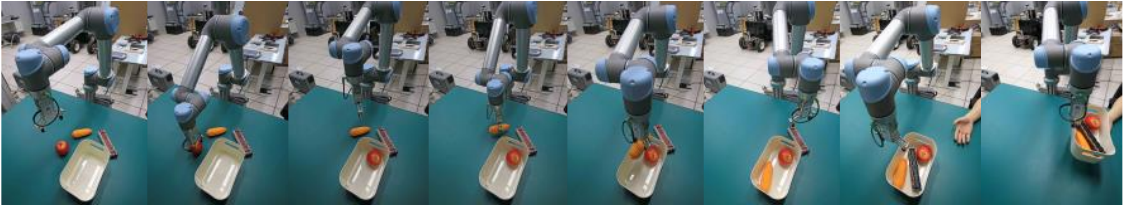}
                \label{multitask2}
        }
        \quad
        \subfigure[]{
                \includegraphics[scale=1.51]{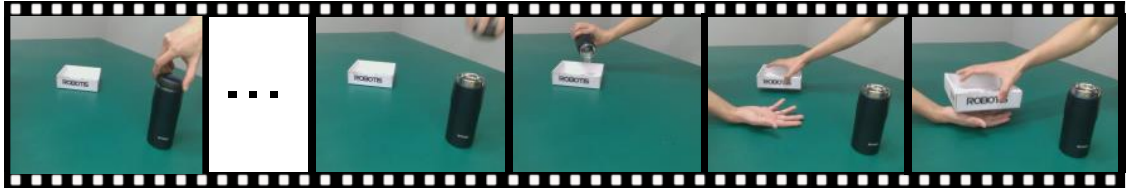}
                \label{multitask3}
        }
        \quad
        \subfigure[]{
                \includegraphics[scale=1.51]{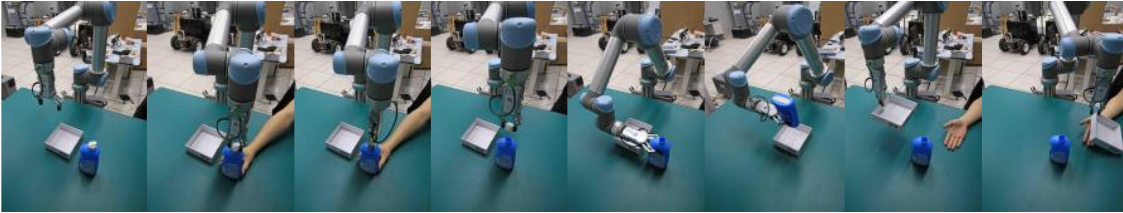}
                \label{multitask4}
        }
        \caption{The examples of robot execution on combined tasks. (a) shows a demonstration of clearing up the table and 
                delivering the white box to human`s hand. (b) shows the robot`s execution of this kind of task in a new environment.
                In subfigure (c), the pictures show a demonstration of openning the bottle, pouring in the box and delivering the box to human`s hand. 
                (d) shows the robot`s execution of this complex task in a new environment.
                } 
        \label{multitasks} 
\end{figure*}
For each task, human demonstration is performed once, while 10 trials of the robot’s task execution are conducted with objects unseen in human demonstration.
 Moreover, to introduce uncertainty in robotic execution, the objects are randomly placed on the work surface for each trial. 
 The example video pictures of robot’s execution on five simple tasks are shown in Fig. \ref{singletasks}, respectively. 
 The success rates and objects used in human demonstration and the robot‘s executions are shown in Table \ref{table-single}. 
 The overall success of our approach across all simple tasks is over $90\%$ in the circumstance of novel objects and changing situations.

\subsection{Complex Tasks Learning}
Furthermore, we test our approach in a more complicated configuration. 
In this setting, two complex tasks composed of two or more simple tasks are designed. 
\begin{table}
        \caption{\textbf{The results of simple tasks}. }
        \label{multiobj-result} % Give a unique label
        \begin{tabular}{cm{2cm}m{1.65cm}c}
        % \hline\noalign{\smallskip}
        \toprule[1pt]
        Task & Demonstration Objects & Workspace\quad\quad\quad Objects & Success Rate\\
        \midrule[1pt]
        pick-place & plate, apple & plastic-box\quad\quad\quad\quad banana & $100\%(10/10)$\\
        \noalign{\smallskip}\hline
        push away & carrot & grape\quad\quad\quad\quad croissant & $100\%(10/10)$\\
        \noalign{\smallskip}\hline
        open bottle & black-bottle \quad\quad\quad plastic-box & paper-box\quad\quad\quad blue-bottle & $90\%(9/10)$\\
        \noalign{\smallskip}\hline
        pour water & paper-box \quad\quad\quad banana \quad\quad\quad black-bottle & paper-box \quad\quad\quad blue-bottle & $90\%(9/10)$\\
        \noalign{\smallskip}\hline
        deliver object & ginger & grape & $100\%(10/10)$\\
        % \noalign{\smallskip}\hline
        \bottomrule[1pt]
        \label{table-single}
        \end{tabular}
        \end{table}
As shown in Fig. \ref{multitasks}(a), in the first complex task, the demonstrator puts all items on the table into a white box and delivers the box to another people. 
In the second complex task, as shown in Fig. \ref{multitasks}(c), the demonstrator grasps and tilts a black bottle to pour something 
into a white box and then delivers the box to another people. 
Video pictures of the robot’s execution are shown in Fig. \ref{multitasks}(b)(d), respectively. 
Similarly to, the robot executes the tasks with novel objects and their changing positions. 
We perform 10 runs for each task and record success rates of the results. 
Success rate and the objects used in human demonstration and the robot’s executions are shown in Table \ref{table-multi}.

 In our experiments, failure cases occur in both simple and complex task execution. 
 Some failures are blamed on object detection, where no object was discovered or where the discovery is wrong or incomplete. 
 The other cases are related to manipulation learner, where the sequence of action primitives extracted is partially inconsistent, 
 possibly leading to conflicting action choices in action planner. 
 In general, these failures could be prevented by increasing training data.

\begin{table}
        \caption{\textbf{The results of complex tasks}. }
        \label{multi-result} % Give a unique label
        \begin{tabular}{cm{3cm}m{1.65cm}c}
        \toprule[1pt]
        Task & Demonstration Objects & Workspace\quad\quad\quad Objects & Success Rate\\
        \midrule[1pt]
        1 & carambol \quad\quad\quad\quad\quad croissant  \quad\quad\quad\quad\quad\quad\quad\quad cake \quad\quad\quad\quad\quad\quad\quad\quad paper-box & plastic-box\quad\quad\quad\quad apple \quad\quad\quad\quad toy-train\quad\quad\quad\quad corn & $90\%(9/10)$\\
        \noalign{\smallskip}\hline
        2 & black-bottle\quad\quad\quad\quad paper-box & blue-bottle\quad\quad\quad\quad paper-box & $90\%(9/10)$\\
        \bottomrule[1pt]
        \label{table-multi}
        \end{tabular}
        \end{table}

\section{CONCLUSIONS}
In this work, we propose a robotic learning system capable of learning manipulation skills from visual observations alone. 
Our approach to imitation learning relies on third-person demonstrations, which makes our approach applicable in more general scenarios. 
To achieve the goal, we present how action recognition and objection detection are combined 
with action planning to give a general solution for robot learning. 
Overall, the results show that our combination of techniques is effective and efficient to generate robot skills. 
Experimental evaluation on complex manipulation tasks from the household domain demonstrates the generalization performance of our framework. 
Thus, we believe that our approach offers a promising way to design autonomous robotic systems with human-like manipulation skills. 
However, failure occurred in our experiments still leave room for improvement, especially for action planning. 
A general and robust action planner is an important area for future work.

% \addtolength{\textheight}{-cm}   % This command serves to balance the column lengths



\section*{ACKNOWLEDGMENT}
This work was supported by State Key Laboratory of Software Development Environment under Grant No SKLSDE-2019ZX-03.

%%%%%%%%%%%%%%%%%%%%%%%%%%%%%%%%%%%%%%%%%%%%%%%%%%%%%%%%%%%%%%%%%%%%%%%%%%%%%%%%

\end{document}